\documentclass{article}
\ifdefined\pdfinclusioncopyfonts \pdfinclusioncopyfonts=1 \fi

\usepackage{microtype}
\usepackage{graphicx}
\usepackage{booktabs}
\usepackage{amsmath,amssymb}
\usepackage{multirow}
\usepackage{xcolor}
\usepackage{xurl}
\usepackage{hyperref}

\usepackage[preprint]{mlsys2025}              %

\usepackage{xspace}
\usepackage[most]{tcolorbox} %

\definecolor{caribbeangreen}{rgb}{0.0, 0.8, 0.6}

  \long\def\doyeon#1{}
  \long\def\syo#1{}

\newcommand{\name}{HERALD\xspace}

\newcommand{\para}[1]{\noindent \textbf{#1.}}

\newcommand{\bcircle}[1]{%
  \tikz[baseline=-0.6ex]{
    \node[circle, fill=black, text=white, minimum size=1.0em, inner sep=0.6pt] {\scriptsize #1};
  }%
}

\newcommand{\wcircle}[1]{%
  \tikz[baseline=-0.6ex]{
    \node[circle, draw=black, fill=white, text=black, minimum size=1.0em, inner sep=0.6pt] {\scriptsize #1};
  }%
}

\usepackage{makecell}
\usepackage{tikz}

\usepackage{dblfloatfix}  %

\usepackage{subcaption}
\usepackage{enumitem}

\usepackage{multirow}
\usepackage{booktabs}
\usepackage{tabularray}   %

\newif\ifwithappendix
\withappendixtrue
\graphicspath{{figures/}}

\mlsystitlerunning{HERALD: Block Diffusion LLM Inference via CPU-GPU KV Retrieval}

\begin{document}

\twocolumn[
\mlsystitle{HERALD: High-Throughput Block Diffusion LLM Inference via CPU-GPU Cooperative KV Cache Retrieval}

\begin{mlsysauthorlist}
\mlsysauthor{Omin Kwon}{snu}
\mlsysauthor{Doyeon Kim}{snu}
\mlsysauthor{Jongseok Park}{berkeley}
\mlsysauthor{Seung Yul Lee}{snu}
\mlsysauthor{Sehoon Kim}{kaist}
\mlsysauthor{Ion Stoica}{berkeley}
\mlsysauthor{Jae W. Lee}{snu}
\end{mlsysauthorlist}

\mlsysaffiliation{snu}{Seoul National University, South Korea}
\mlsysaffiliation{berkeley}{University of California, Berkeley, USA}
\mlsysaffiliation{kaist}{KAIST, South Korea}
\mlsyscorrespondingauthor{Jae W. Lee}{jaewlee@snu.ac.kr}

\mlsyskeywords{Block Diffusion LLMs, KV Cache Offloading, Sparse Attention, CPU-GPU Cooperative Inference, LLM Inference}

\vskip 0.3in

\begin{abstract}
Block diffusion LLMs (block dLLMs) improve decode throughput by generating tokens in parallel, and are emerging as a next-generation alternative to autoregressive LLMs. 
At long contexts, however, the KV cache remains a major throughput bottleneck in both memory bandwidth and capacity. 
For autoregressive LLMs, dynamic sparse attention is often combined with KV offloading to alleviate both bottlenecks, keeping the full KV cache in CPU DRAM and fetching only a critical subset to the GPU. 
However, KV subset selection and retrieval over bandwidth-limited PCIe incur substantial offloading overheads that often stall the GPU. 
We present HERALD, the first dynamic sparse attention method for block dLLMs that supports KV offloading. 
HERALD exploits a property of block dLLMs: a critical KV subset selected once per block is reused across all of its denoising steps, so the next block’s selection can be hidden behind the current block’s denoising. 
HERALD realizes this with a two-stage selection algorithm co-designed with its execution system. 
While the GPU denoises the current block, the CPU speculatively searches the full KV cache for the next block with a single query merged from the current block’s B queries, cutting the selection compute by B$\times$; the GPU then refines the candidates into the final top-$K$ subset with the next block’s real queries. At $K{=}1024$, HERALD achieves near-lossless accuracy relative to Dense on LongBench and reasoning tasks across four block dLLMs, and improves decode throughput by up to 3.70$\times$ over Dense GPU-only inference, 2.59$\times$ over MAGE, a state-of-the-art block-dLLM sparse attention method, and 8.38$\times$ over MAGE-Offload, its direct KV-offloading extension.

\end{abstract}
]

\printAffiliationsAndNotice{}

\section{Introduction}
\label{sec:intro}

Diffusion LLMs (dLLMs) have been proposed as an alternative to autoregressive LLMs (AR LLMs), which generate one token per forward pass during decoding and thus leave GPU compute underutilized~\cite{11122346,leviathan2023fastinferencetransformersspeculative}.
By denoising multiple tokens in parallel, dLLMs can produce several tokens per forward pass. Early dLLMs, however, attend bidirectionally over the entire sequence~\cite{nie2025large, ye2025dream7b}, so they lack an exact KV cache and must recompute attention over all tokens at every denoising step~\cite{wu2025fastdllmtrainingfreeaccelerationdiffusion}.
Block diffusion LLMs (block dLLMs) resolve this by interpolating between the two paradigms. They generate blocks of a sequence autoregressively, as AR LLMs generate tokens, while denoising the tokens within each block in parallel, as dLLMs do~\cite{arriola2025blockdiffusioninterpolatingautoregressive, wu2025fastdllmv2efficientblockdiffusion}.
Specifically, the $B$ tokens of a block are decoded in parallel over $T$ denoising steps, and once a block is completed, its KV cache is reused by subsequent blocks.
These block dLLMs match the accuracy of AR models at the same scale~\cite{bie2025llada20scalingdiffusionlanguage,cheng2025sdarsynergisticdiffusionautoregressionparadigm}. The block-wise design has become the dominant form of dLLMs, adopted by industry models including ByteDance's Seed Diffusion~\cite{song2025seeddiffusionlargescalediffusion}, Google's DiffusionGemma~\cite{diffusiongemmateam2026diffusiongemmatechnicalreport}, and NVIDIA's Nemotron-Labs Diffusion~\cite{fu2026nemotronlabsdiffusiontrimodelanguagemodel}, and supported natively by serving engines such as SGLang~\cite{zheng2024sglangefficientexecutionstructured} and vLLM~\cite{kwon2023efficientmemorymanagementlarge}, making block dLLMs a practical next-generation alternative to AR LLMs.

\begin{figure}[t]
  \centering
  \includegraphics[width=\columnwidth]{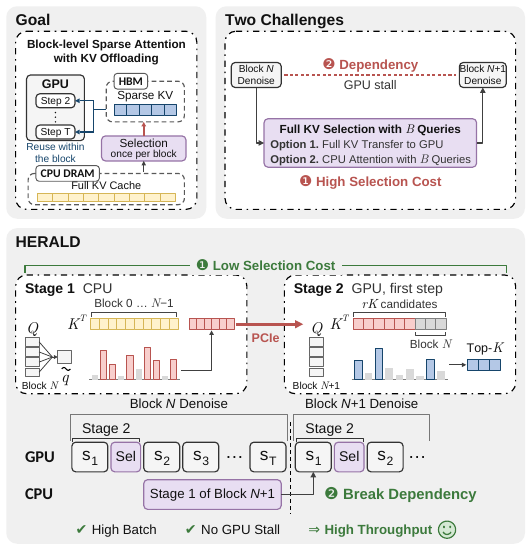}
  \caption{
  Overview of \name. Applying sparse attention with KV offloading to
  block dLLMs to improve throughput faces two challenges:
  \bcircle{1} selecting each block's critical KV subset
  requires access to the full cache, which is costly under offloading,
  and \bcircle{2} the sequential dependency between selection and
  denoising causes GPU stalls. \name addresses both with two-stage
  selection: Stage~1 cheaply searches for candidates on the CPU while
  the GPU denoises block $N$, and Stage~2 selects the critical KV subset
  for block $N{+}1$ from these candidates on the GPU.
  }
  \label{fig:intro}
\end{figure}

Block dLLMs thus recover the compute efficiency that AR decoding lacks, but because they also rely on the KV cache, they inherit the two memory bottlenecks of AR LLMs, which worsen with context length and lower throughput.
Every denoising step reads the KV cache of all completed blocks, creating a \emph{memory-bandwidth} bottleneck.
Storing the KV cache in GPU HBM also creates a \emph{memory-capacity} bottleneck that limits batch size and throughput.
To relieve both bottlenecks, KV eviction methods~\cite{song2025sparsedllmacceleratingdiffusionllms, huang2025maskkv} discard KV entries judged unimportant.
However, evicted entries can never be attended to again, so these methods lose accuracy on long-generation tasks.

Dynamic sparse attention~\cite{wang2025sparsedsparseattentiondiffusion, xi2026losa, kwon2026mage} instead retains the full KV cache and dynamically selects only the critical KV subset needed for attention, alleviating the memory-bandwidth bottleneck while preserving accuracy.
Combining it with KV offloading could also alleviate the memory-capacity bottleneck. The full KV cache resides in CPU DRAM, whose capacity is an order of magnitude larger than HBM, and only the selected subset is fetched to the GPU, enabling larger batches. This, however, comes with a substantial offloading overhead. Selecting the critical KV subset and transferring it from CPU to GPU over bandwidth-limited PCIe both take time, and since GPU attention cannot start until the selected KV entries arrive, this overhead adds directly to execution time.
Mitigating this overhead is a key challenge in realizing
the throughput benefits of KV offloading, motivating many sparse attention methods for AR LLMs to reduce it or hide it behind GPU
computation~\cite{lee2024infinigenefficientgenerativeinference,
sun2025shadowkvkvcacheshadows,
liu2026freekvboostingkvcache,
xu2025specontextenablingefficientlongcontext,
liu2025clusterkvmanipulatingllmkv,
wu2025louiskvefficientkvcache, flexicache, opkv}.

In this work, we exploit a unique opportunity that exists in block
dLLMs to mitigate offloading overhead: block-level sparse attention. MAGE~\cite{kwon2026mage}, a state-of-the-art dynamic sparse attention method for block dLLMs, observes that the critical KV subset changes little within a block, so it selects the subset once per block using the block's first-step \texttt{[MASK]} queries and reuses it for the remaining steps. Under KV offloading, this block-level selection amortizes both selection and CPU-to-GPU retrieval costs over the block's $T$ denoising steps, an advantage unavailable to AR LLMs.
This reduced selection frequency alone, however, does not make KV offloading efficient for block dLLMs. Simply offloading MAGE, which we call \emph{MAGE-Offload}, faces two system challenges (Figure~\ref{fig:intro}, top right). Its selection requires costly attention between the block's $B$ queries and the full KV cache, and it still depends on the previous block's completion, which keeps it on the critical path. As a result, MAGE-Offload reaches only 0.27--0.52$\times$ the throughput of GPU-only inference despite its larger batches (Figure~\ref{fig:throughput}).

To address these challenges, we present \name, to our
knowledge the first dynamic sparse attention method for
block dLLMs that supports KV offloading.
\name co-designs the selection algorithm and execution system
around two-stage selection (Figure~\ref{fig:intro}, bottom).
Stage~1 approximates MAGE's selection to address both system
challenges: it uses the previous block's first-step queries
as a proxy to break the sequential dependency and merges
these queries into one to reduce CPU selection cost.
Together, these approximations allow CPU selection for the
next block to overlap with GPU denoising of the current block.
To compensate for approximation errors, Stage~1 selects an
expanded candidate set rather than the final critical KV subset.
Stage~2 then refines these candidates on the GPU using the
target block's real queries, recovering selection quality.

We implement \name on SGLang~\cite{zheng2024sglangefficientexecutionstructured}
and evaluate it across four block dLLMs on
LongBench~\cite{bai2024longbenchbilingualmultitaskbenchmark}
and long-generation reasoning benchmarks,
MATH-500~\cite{math500} and LiveCodeBench~\cite{lcb}.
We make the following contributions:
{\setlength{\parskip}{0pt}
\begin{itemize}[
  leftmargin=*,
  topsep=5pt,
  partopsep=0pt,
  itemsep=7pt,
  parsep=0pt
]
 \item We propose \name, the first sparse attention system for block dLLMs that supports KV offloading, exploiting block-level KV reuse to amortize offloading overheads (\S\ref{sec:herald}).
 \item We co-design a \name algorithm and
 its execution system that realizes this opportunity: the CPU speculatively searches for
 KV candidates at low cost, while the GPU selects the final
 top-$K$ subset and denoises with sparse attention
 (\S\ref{sec:herald:core-design}, \S\ref{sec:lookahead}).   
  \item \name achieves near-lossless accuracy relative to Dense and up to
  3.70$\times$ the decode throughput of GPU-only inference and
  8.38$\times$ that of MAGE-Offload (\S\ref{sec:eval}).
\end{itemize}
}

\section{Related Work}
\label{sec:background}

\subsection{Block Diffusion LLM Inference}
\label{sec:bg:block-dllm}
\begin{figure}[t]
  \centering
  \includegraphics[width=\columnwidth]{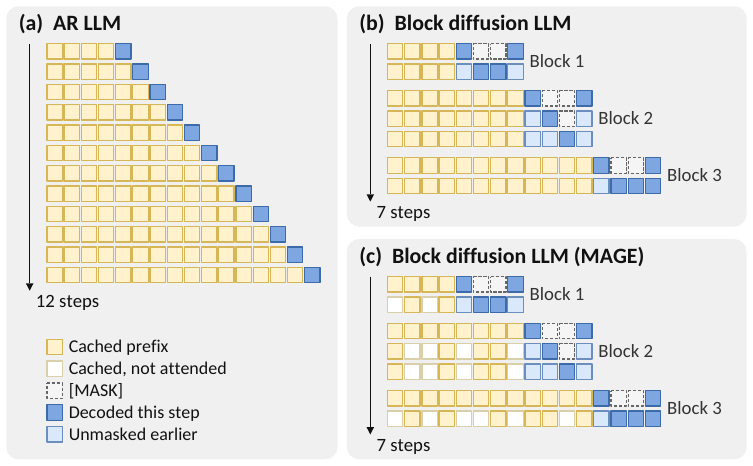}
  \caption{Decoding in (a) an AR LLM, (b) a block diffusion LLM ($B{=}4$), and (c) MAGE, a sparse attention method for long-context acceleration.}
  \label{fig:diffusion-process}
  \vspace{-8pt}
\end{figure}

\para{From AR LLMs to block dLLMs}
Autoregressive (AR) LLMs generate one token per sequence with each
forward pass, which reads the model weights and the prefix KV cache for a
single new token and leaves GPU compute
underutilized~\cite{11122346, leviathan2023fastinferencetransformersspeculative,
Stojkovic_2025}.
Diffusion LLMs (dLLMs) relax this token-level dependency by jointly
refining multiple token positions over a sequence of denoising steps
~\cite{gong2023diffuseq,
sahoo2024simpleeffectivemaskeddiffusion,
lou2024discretediffusionmodelingestimating,
nie2025large,
ye2025dream7b,
gong2025scalingdiffusion}.
Early dLLMs, however, use full-sequence bidirectional attention at every
step, preventing exact KV caching and requiring repeated processing of the
entire sequence, which approximate caching methods only partially
mitigate~\cite{liu2025dllmcacheacceleratingdiffusionlarge,
ma2025dkvcachecachediffusionlanguage}.
Block diffusion models instead support exact KV caching by
generating fixed-size blocks of $B$ tokens autoregressively,
with diffusion-based parallel decoding within each block~\cite{arriola2025blockdiffusioninterpolatingautoregressive,
wu2025fastdllmv2efficientblockdiffusion,
kim2026cdlmconsistencydiffusionlanguage,
bie2025llada20scalingdiffusionlanguage,
cheng2025sdarsynergisticdiffusionautoregressionparadigm}.
With higher throughput and AR-level accuracy, block diffusion
has become the dominant dLLM approach, adopted by open and
industry-developed models~\cite{bie2025llada20scalingdiffusionlanguage,
cheng2025sdarsynergisticdiffusionautoregressionparadigm,
wu2026dreamreasoner8bblocksizecurriculumlearning,
gemini_diffusion,
song2025seeddiffusionlargescalediffusion,
fu2026nemotronlabsdiffusiontrimodelanguagemodel}.

\para{Block dLLM inference}
As Figure~\ref{fig:diffusion-process}(b) shows, each block
starts as $B$ \texttt{[MASK]} tokens and undergoes $T$
denoising steps, attending bidirectionally within the block
and causally to completed blocks.
At each step, the scheduler accepts predictions at selected
masked positions~\cite{wu2025fastdllmtrainingfreeaccelerationdiffusion,
kim2025trainworstplanbest}.
Once all tokens are accepted, the block's KV states are
appended to the prefix cache.
Thus, the prefix KV cache stays fixed within a block,
while its queries change as tokens are accepted.

\subsection{Long-Context Block dLLM Acceleration}
\label{sec:bg:acceleration}
\label{sec:bg:offloading}

At long contexts, the prefix KV cache creates two memory bottlenecks that
lower the throughput.
Reading the KV cache for attention dominates each denoising step, a
\emph{memory-bandwidth} bottleneck, and storing the KV cache occupies most
of GPU HBM, a \emph{memory-capacity} bottleneck that caps the batch size.
Prior work addresses these bottlenecks in two ways.

\para{KV eviction}
KV eviction addresses both bottlenecks by discarding the entries judged
unimportant.
Sparse-dLLM~\cite{song2025sparsedllmacceleratingdiffusionllms} evicts
low-saliency entries with a delay, and MaskKV~\cite{huang2025maskkv}
scores prompt tokens with the attention of \texttt{[MASK]} queries.
The entries that matter, however, shift as generation proceeds, and an
evicted entry can never be attended to again, so accuracy degrades on
long-generation tasks.

\para{Dynamic sparse attention}
Dynamic sparse attention instead keeps every entry and dynamically selects
the critical KV subset that each step needs, which relieves the
memory-bandwidth bottleneck.
SparseD~\cite{wang2025sparsedsparseattentiondiffusion}, for fully
bidirectional dLLMs, derives head-specific sparse patterns from full
attention in early steps and reuses them.
For block dLLMs, LoSA~\cite{xi2026losa} applies sparse attention only to
the tokens whose representations still change.
MAGE also targets block dLLMs but selects sparsity at the block level: as shown in Figure~\ref{fig:diffusion-process}(c), it selects
the critical KV subset once per block using exact attention
between the block's first-step \texttt{[MASK]} queries and
the full KV cache. It reuses this subset for the remaining
$T-1$ denoising steps within that block, reducing KV accesses.
The memory-capacity bottleneck, however, remains: these methods keep the
full KV cache in GPU.
AR LLMs face the same bottleneck, and their sparse attention methods
remove it with KV offloading, keeping the full KV cache in CPU DRAM and
fetching only the selected subset to the
GPU~\cite{lee2024infinigenefficientgenerativeinference,
sun2025shadowkvkvcacheshadows,
wu2025louiskvefficientkvcache,
liu2025clusterkvmanipulatingllmkv,
xu2025specontextenablingefficientlongcontext,
liu2026freekvboostingkvcache,
flexicache,opkv}.
To our knowledge, no sparse attention method for block dLLMs supports KV
offloading, and the existing AR methods supporting KV offloading degrade
accuracy on block dLLMs, as their estimation of the critical KV subset is
tailored to AR decoding.

\section{Motivation}
\label{sec:motivation}

This section shows why long-context block dLLM inference needs KV
offloading (Section~\ref{sec:motivation:capacity}), states our goal of
block-level sparse attention over an offloaded KV cache
(Section~\ref{sec:motivation:goal}), and explains why this goal is
challenging (Sections~\ref{sec:motivation:problem}
and~\ref{sec:motivation:dependency}).
Unless stated otherwise, the motivation experiments use
LLaDA2.0-mini~\cite{bie2025llada20scalingdiffusionlanguage} with a 32K-token
context, block size $B{=}32$, and $T{=}20$ denoising steps on the
platform of Section~\ref{sec:setup}.

\subsection{Problem: KV Cache Capacity Pressure Limits Throughput}
\label{sec:motivation:capacity}

\begin{figure}[t]
  \centering
  \includegraphics[width=\columnwidth]{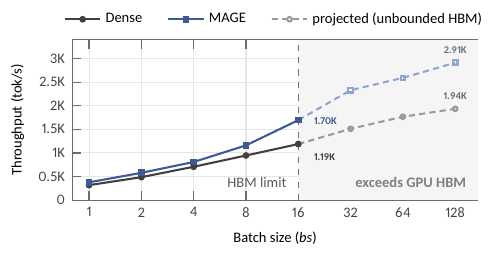}
  \caption{Decode throughput of Dense and MAGE with the full KV cache in
  HBM as the batch size grows (LLaDA2.0-mini, 32K context, H100). HBM runs
  out at \textit{bs}${=}16$, well before throughput saturates; larger
  batches are projected assuming unbounded HBM.}
  \label{fig:motivation-ceiling-a}
\end{figure}

Block dLLMs decode a block of $B$ tokens per forward pass, but their KV
cache still grows linearly with the context: each completed block appends
its KV entries, exactly as $B$ tokens of an AR LLM would.
At long contexts the KV cache therefore occupies most of the HBM and caps
the batch size well before the GPU's compute saturates.
Figure~\ref{fig:motivation-ceiling-a} shows this on LLaDA2.0-mini at
32K: Dense runs out of HBM at a batch size of 16 and 1.19K tokens/s,
short of the 1.94K tokens/s it would reach at a batch size of 128 with
unlimited HBM. MAGE hits the same limit at 1.70K tokens/s, short of 2.91K
tokens/s.
Offloading the full KV cache to CPU DRAM, whose capacity is an order of
magnitude larger, is therefore a prerequisite for high-throughput inference.

\subsection{Goal: Block-Level Sparse Attention under KV Offloading}
\label{sec:motivation:goal}

To relieve this capacity pressure, block dLLMs need a sparse attention
method that supports KV offloading: the full KV cache resides in CPU DRAM,
and only the critical KV subset is fetched to the GPU.
However, selecting this subset and transferring it over PCIe
introduce substantial overheads that can limit the throughput
gains of offloading~\cite{
sun2025shadowkvkvcacheshadows,
xu2025specontextenablingefficientlongcontext,
liu2026freekvboostingkvcache, flexicache, opkv}.

To overcome these overheads, we build on the selection insight
of MAGE~\cite{kwon2026mage}, the state-of-the-art dynamic sparse
attention method for block dLLMs.
MAGE selects the critical KV subset at the first step of
each block and reuses it for the remaining denoising steps.
Under KV offloading, this reuse amortizes both selection
and CPU-to-GPU retrieval over the block's $T$ denoising steps.

\begin{figure}[t]
  \centering
  \includegraphics[width=\columnwidth]{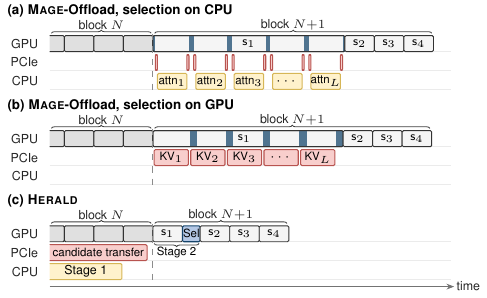}
  \caption{Pipeline diagram of MAGE-Offload and \name.
  (a) MAGE-Offload with CPU selection over the full KV cache.
  (b) MAGE-Offload with GPU selection over the full KV cache.
  (c) \name performs selection in two stages:
  \hyperref[sec:herald:stage1]{Stage~1} that speculatively searches
  for candidates on CPU and \hyperref[sec:herald:stage2]{Stage~2}
  that narrows down the candidates on GPU, thereby minimizing the selection overhead on critical path.
  }
  \label{fig:motivation-dependency}
\end{figure}

The most direct design, which we call \emph{MAGE-Offload}, applies the
MAGE algorithm directly to the offloaded KV cache.
There are two options to adapt the MAGE algorithm to the offloading
design, which differ in where the selection runs.
In the first option, the CPU performs the selection by computing the
attention between the queries of the block and the full KV cache in CPU
DRAM (Figure~\ref{fig:motivation-dependency}(a)).
In the second option, the GPU performs the selection after streaming the
full KV cache from CPU DRAM over PCIe
(Figure~\ref{fig:motivation-dependency}(b)).
Either way, MAGE-Offload faces two challenges, which the next two
subsections detail.

\subsection{Challenge 1: High Cost of Critical KV Selection under Offloaded Full KV}
\label{sec:motivation:problem}
\label{sec:motivation:challenges}

\begin{figure}[t]
  \centering
  \includegraphics[width=\columnwidth]{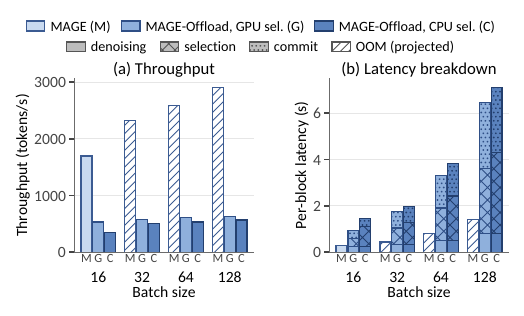}
  \caption{(a) Decode throughput of MAGE and MAGE-Offload across batch
  sizes and (b) their per-block latency breakdown (LLaDA2.0-mini, 32K
  context, $K{=}1024$, $T{=}20$). MAGE-Offload sustains larger batches but
  its throughput stays far below MAGE, as the exposed critical KV selection
  and the commit of the block's KV cache dominate its per-block latency.}
  \label{fig:motivation-cost}
\end{figure}

MAGE selects the critical KV subset of a block at its first step, where the
block's $B$ \texttt{[MASK]} queries attend to the full prefix KV cache.
With the full KV cache offloaded to CPU DRAM, this selection can run on
either the GPU or the CPU, and both are costly.
In addition, once the block is fully denoised, the forward that computes
the block's final KV cache also attends to the full KV cache.
As Figure~\ref{fig:motivation-cost} shows, the selection and this commit
forward inflate the per-block latency so much that, despite the larger
batches that offloading enables, MAGE-Offload ends up with lower
throughput than GPU-only inference.

\para{Selection on the GPU}
To select with exact attention over the full KV cache on the GPU,
MAGE-Offload must stream the cache to the GPU layer by layer over PCIe,
whose bandwidth is two orders of magnitude below HBM.
This transfer takes far longer than the block's denoising
(Figure~\ref{fig:motivation-cost}(b)).
The commit forward streams the full KV cache over PCIe once more, so the
commit costs about as much as the selection.

\para{Selection on the CPU}
Selecting on the CPU avoids the full-KV transfer over PCIe, but CPU
memory bandwidth is an order of magnitude lower than GPU HBM, so reading
the full KV cache already takes longer.
Worse, block dLLMs make the selection compute-bound: all $B$
\texttt{[MASK]} queries of a block attend to the full KV cache at once,
raising the arithmetic intensity of attention by $B{\times}$ over AR
decoding with GQA, beyond the ridge point of CPUs
(Appendix~\ref{app:roofline}), whose compute is far below the GPU's.
As a result, the selection on the CPU takes even longer than the
selection on the GPU, on top of the same commit cost
(Figure~\ref{fig:motivation-cost}(b)).

\begin{figure*}[t]
  \centering
  \includegraphics[width=\textwidth]{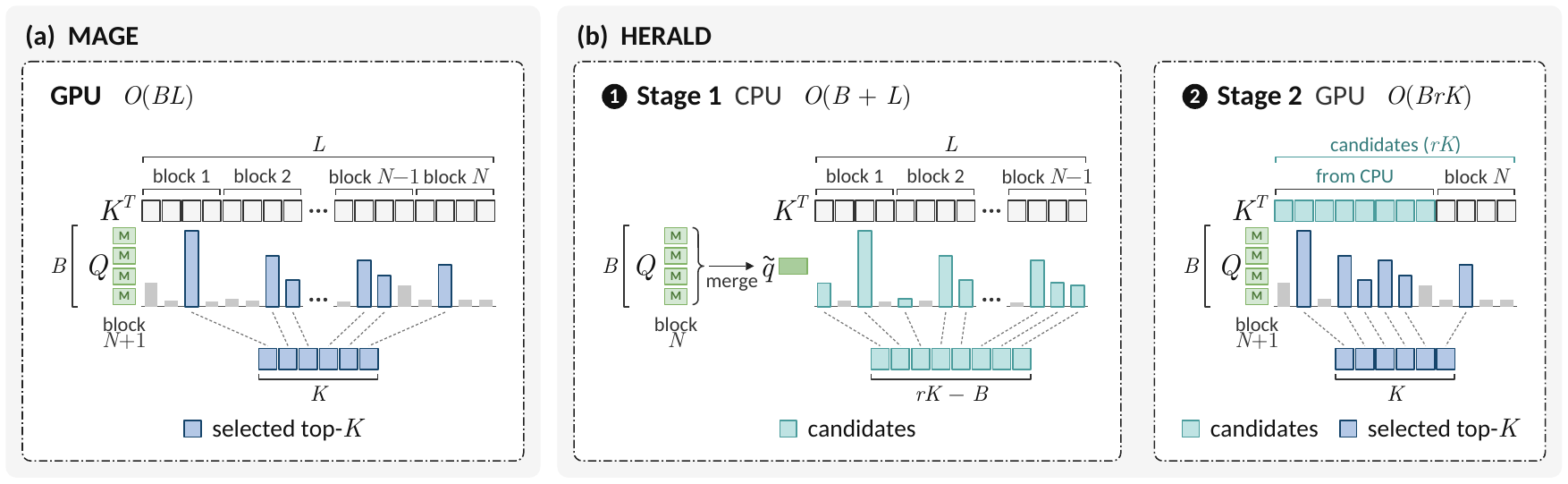}
  \caption{Algorithm overview of \name.
  (a) KV selection in MAGE. At the first step of each block, the $B$
  \texttt{[MASK]} queries $Q$ score the full KV cache via exact attention
  and select the top-$K$ critical KV subset.
  (b) KV selection in \name proceeds in two stages. In Stage~1, the CPU
  speculatively searches the full KV cache for candidates, using the
  previous block's $B$ \texttt{[MASK]} queries as a proxy. These queries
  are merged into a single query to reduce CPU computation.
  In Stage~2, at the first step of the current block, the real
  \texttt{[MASK]} queries score the candidate set and select the top-$K$
  critical KV subset, following MAGE's selection rule.}
  \label{fig:selection-stages}
  \vspace{-0.3cm}
\end{figure*}

\subsection{Challenge 2: Sequential Dependency of Critical KV Selection and Denoising}
\label{sec:motivation:dependency}

The high selection cost is aggravated because the selection sits on the
critical path of every block.
Block $N{+}1$'s selection runs at its first step, where its \texttt{[MASK]}
queries attend to the full prefix KV cache, including block $N$, to pick
the critical KV subset.
Block $N$'s KV cache, however, exists only after block $N$'s denoising
completes, and block $N{+}1$'s remaining steps need the selected critical KV subset.
Selection on either device therefore stalls the GPU at every block
boundary (Figure~\ref{fig:motivation-dependency}(a) and (b)).
Selection and denoising thus alternate strictly, and the selection
cost is exposed in every block.
Hiding the selection requires starting block $N{+}1$'s selection before
block $N$ completes, which this dependency rules out.

\section{\name}\label{sec:herald}

\name is a sparse attention method for block dLLMs that offloads
the full KV cache to CPU DRAM for high-throughput inference with long contexts.
It builds on MAGE~\cite{kwon2026mage}'s block-level selection, which identifies a
high-quality critical KV subset at each block's first step through
attention between the $B$ \texttt{[MASK]} queries and the full KV cache.
Directly applying this selection algorithm under KV offloading,
however, severely degrades system performance, as discussed in
Sections~\ref{sec:motivation:challenges}
and~\ref{sec:motivation:dependency}.
We therefore co-design the critical KV selection algorithm and its
execution system to support efficient sparse attention under KV offloading.
Section~\ref{sec:herald:core-design} presents the algorithm design,
and Section~\ref{sec:lookahead} presents the system design.

\subsection{HERALD Algorithm Design}
\label{sec:herald:core-design}

\para{Algorithm overview}
MAGE selects, for each KV head, the $K$ entries with the highest attention scores in the full KV cache as the block's critical KV subset, where $K$ is the number of entries retained per head.
As Figure~\ref{fig:selection-stages} shows, \name splits this selection  into two stages: candidate search on the CPU (Stage~1) and final
selection on the GPU (Stage~2).
Stage~1 resolves the system challenges described in Sections~\ref{sec:motivation:challenges}
and~\ref{sec:motivation:dependency} by approximating MAGE's selection,
exploiting the properties of block dLLMs.
First, it lowers the CPU selection cost by merging the $B$ \texttt{[MASK]} queries into a single query, reducing the attention computation, which is heavily compute-bound on the CPU, by a factor of $B$.
Second, it uses the current block $N$'s first-step queries as a proxy
for the next block $N{+}1$'s first-step queries for its selection.
Thus, selection for the next block $N+1$ can begin immediately after the current block $N$'s first step completes, without waiting for
block $N$ to be fully generated, breaking the sequential dependency.
To compensate for approximation errors, Stage~1 forms an expanded candidate KV set containing $rK$ entries per KV head, where $r$
specifies how many times larger the candidate set is than the final critical KV subset of $K$ entries.
Stage~2 re-ranks these candidates on the GPU and selects the final $K$ entries of critical KV set, following MAGE's selection rule. 
This two-stage selection algorithm allows the CPU candidate search for block $N{+}1$ to \textit{completely overlap} with the GPU's remaining denoising steps for block $N$, enabling high-quality, high-throughput sparse attention under KV offloading. We describe Stage~1 and Stage~2 in detail below.

\phantomsection
\label{sec:herald:stage1}\label{sec:herald:representative-query}\label{sec:herald:draft}
\para{Stage 1: candidate search with approximations on the CPU}
Stage~1 exploits the block dLLM property that, at the first step of each block, all query positions contain the same \texttt{[MASK]} token;
only their token positions differ.
This structure suggests that the queries of the first step which are used for selection can be efficiently approximated to address the two system challenges while preserving useful information for KV selection.
Stage~1 uses two such approximations: using the previous block's queries
as a proxy to break the sequential dependency and merging the queries
to reduce CPU selection cost.

First, the first-step queries of block $N{+}1$ exist only after block $N$ completes,
so Stage~1 of block $N+1$ cannot start until the GPU produces block $N$.
Thus, \name uses the queries of block $N$ as a
proxy for those of block $N{+}1$.
Specifically, it uses the first-step queries of block $N$, which the GPU has
already computed at the first step of block $N$, so the proxy costs no extra
forward pass.
With these queries, the CPU speculatively searches the KV cache
of blocks $0$ to $N{-}1$ for candidates for block $N{+}1$ while
the GPU runs the remaining denoising steps of block $N$.
Once block $N$ completes, its $B$ KV entries are added to the
candidate set.

\begin{figure*}[t]
  \centering
  \includegraphics[width=\textwidth]{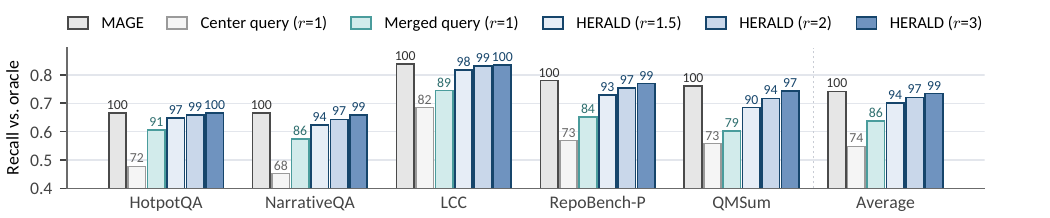}
  \caption{Selection quality of \name's two-stage selection compared with
  MAGE, measured as recall against the
  per-step oracle---the exact top-$K$ under the real queries at each
  denoising step.
  Center query and merged query are ablations without Stage~2: they
  directly select $K$ entries ($r{=}1$) using the block's center-position
  query and the mean of its queries, respectively.
  \name achieves 94\%, 97\%, and 99\% of MAGE's recall
  at $r{=}1.5$, $2$, and $3$, respectively, averaged over five LongBench
  tasks. Numbers above the bars indicate percentages of MAGE's recall.
  (LLaDA2.0-mini, $K{=}1024$.)}
  \label{fig:recall-recovery}
  \vspace{-0.3cm}
\end{figure*}

Second, 
MAGE ranks each prefix key by its attention probability averaged over
the $B$ queries of the block, i.e., the mean of softmax.
Stage~1 of HERALD instead first merges the $B$ queries into one and ranks the keys by the
softmax of the mean:
\begin{equation}
  \scalebox{0.85}{$
    \begin{array}{@{}l@{\quad}l@{}}
      \text{MAGE:}
      & \frac{1}{B}\sum_{i=1}^{B}
        \mathrm{softmax}\bigl(\mathbf{q}_i K^{\top}\bigr), \\[3pt]
      \text{HERALD Stage 1:}
      & \mathrm{softmax}\!\left(
        \left(\frac{1}{B}\sum_{i=1}^{B}\mathbf{q}_i\right)K^{\top}
        \right).
    \end{array}
  $}
  \label{eq:merged-query}
\end{equation}
where $\mathbf{q}_i$ is the first-step query at block position $i$, $K$
holds the prefix keys, and the mean of the queries is the merged query.
As shown in Figure~\ref{fig:selection-stages}, the per-head scoring cost
is reduced from $\mathcal{O}(BL)$ for MAGE to $\mathcal{O}(B+L)$ for
HERALD's Stage~1, where $L$ is the length of the full KV cache, reducing the
computational complexity of selection on the CPU.
Figure~\ref{fig:recall-recovery} demonstrates the effectiveness of
our single-query approximation (Equation~\ref{eq:merged-query}).
Merging the $B$ queries achieves higher recall than another
single-query approach that uses only the query at the center
position of the block. 
However, directly selecting the critical KV subset using the merged query achieves only 86\% of MAGE's recall on average across the five LongBench tasks.

To compensate for ranking errors from both query merging and
using the previous block's queries as a proxy, \name forms an
expanded candidate set of $rK$ entries rather than directly
selecting the final $K$ entries.
Stage~2 then selects the final $K$ entries from this candidate set.

\phantomsection
\label{sec:herald:stage2}
\para{Stage 2: final selection over the candidates on the GPU}
Stage~2 selects the critical KV subset among the $rK$ candidates on the
GPU with the first-step queries of block $N{+}1$, following MAGE's rule.
At the first step of block $N{+}1$, the $B$ queries attend to the candidates,
and the top-$K$ entries per KV head are kept for the remaining $T{-}1$
steps.
The two stages reproduce MAGE exactly whenever the candidates contain
MAGE's top-$K$ entries, so the only loss comes from MAGE's entries that
Stage~1 misses, which shrinks as $r$ grows.
Stage~2 on GPU requires only $rK/L$ of MAGE's attention-scoring
computation, as it scores the candidates rather than the full
KV cache. The additional computation introduced by \name is
the CPU candidate search in Stage~1. By design, this search
runs speculatively and overlaps with GPU denoising, adding
negligible overhead to the critical path.
Section~\ref{sec:lookahead} details the execution system
that enables this overlap.

\para{Selection quality of \name relative to MAGE}
Beyond reducing MAGE's computational cost, \name recovers
most of the selection quality lost through the Stage~1 approximations by its two-stage design.
Figure~\ref{fig:recall-recovery} shows how closely \name's selection
approaches MAGE's on the evaluation tasks at $K{=}1024$, measured as
recall against the per-step oracle, i.e., the exact top-$K$ critical KV
subset under the real queries at each denoising step.
Stage~2 recovers most of the gap between MAGE and \name even
at small values of $r$, reaching 94\%, 97\%, and 99\% of MAGE's
recall on average at $r{=}1.5$, $2$, and $3$, respectively.
Section~\ref{sec:eval:accuracy} shows that $r{=}2$ largely preserves
downstream accuracy relative to MAGE.
\label{sec:herald:system-gain}

\subsection{HERALD System Design}
\label{sec:lookahead}

\begin{figure*}[t!]
  \centering
  \includegraphics[width=\textwidth]{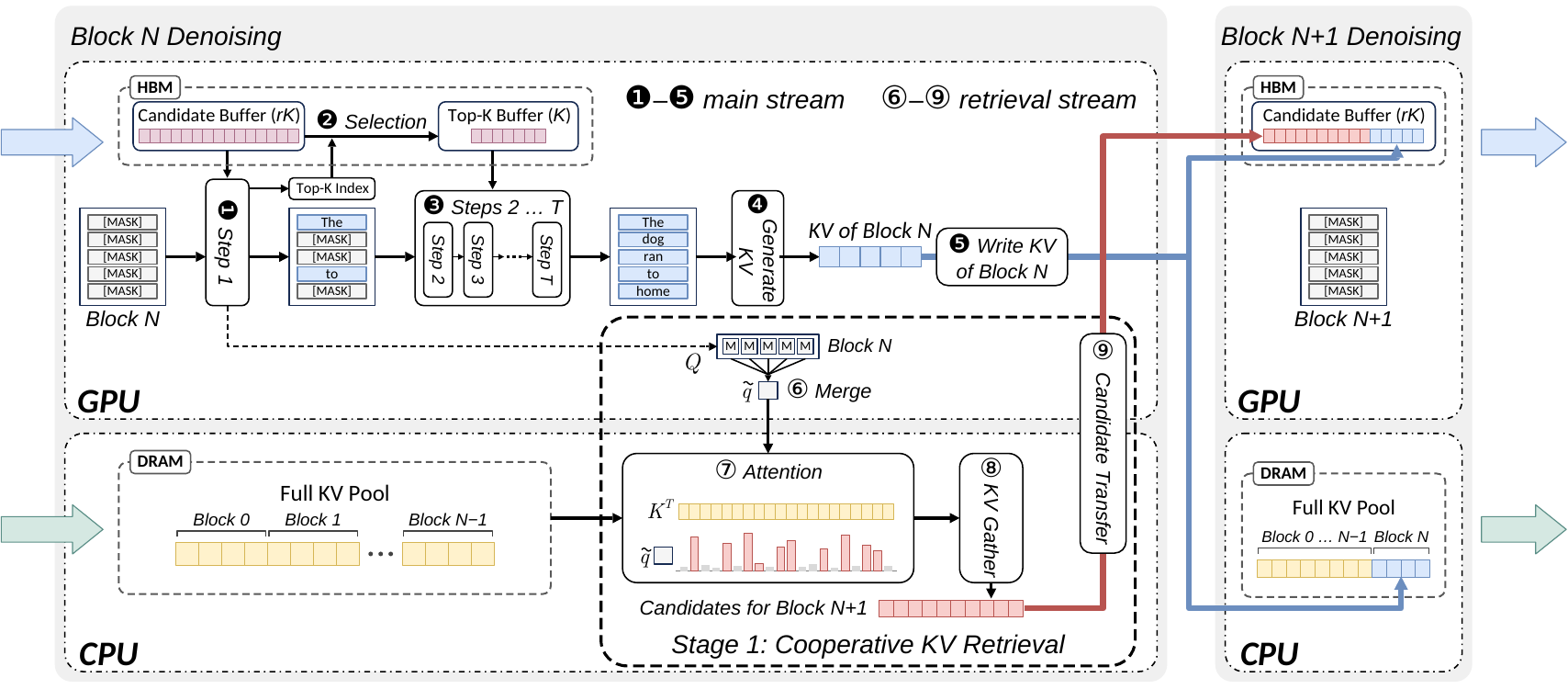}
  \caption{System overview of \name. The main stream (\bcircle{1}--\bcircle{5})
  denoises block $N$ on the GPU. The first step (\bcircle{1}) attends to the $rK$ entries
  of the candidate KV buffer, and Stage~2 uses this attention to select $K$ of them
  (\bcircle{2}) into the top-$K$ KV buffer for steps $2$ to $T$ (\bcircle{3}); the KV cache of the completed
  block (\bcircle{4}) is written to both memories (\bcircle{5}). The retrieval
  stream (\wcircle{6}--\wcircle{9}) runs Stage~1 for block $N{+}1$:
  the GPU merges the $B$ first-step queries of block $N$ into one query
  (\wcircle{6}); the CPU computes attention between the merged query and the
  full KV cache in the DRAM-resident full KV pool
  (\wcircle{7}) and gathers the top-$(rK{-}B)$ entries (\wcircle{8}), which are
  copied into the candidate KV buffer in HBM, next to block $N$'s KV cache
  (\wcircle{9}).}
  \vspace{-0.3cm}
  \label{fig:architecture}
\end{figure*}

\para{System overview}
\label{sec:herald:execution-flow}
As shown in Figure~\ref{fig:architecture}, \name stores the full
KV cache in a CPU-resident full KV pool and maintains two sparse
buffers on the GPU: a candidate KV buffer with $rK$ entries and
a top-$K$ KV buffer with $K$ entries per KV head.
After prefill, \name retains only the top-$K$ KV entries of input tokens on the
GPU and offloads the full KV cache to CPU DRAM. To select these
entries, it includes the first decode block of $B$ \texttt{[MASK]}
tokens in prefill and uses their attention scores.
During decoding, two concurrent streams execute the two-stage
selection algorithm. The main stream denoises block $N$ on the
GPU, performing Stage~2 at its first step. Meanwhile, the retrieval
stream prepares the candidates for block $N{+}1$ through Stage~1
on the CPU and transfers them to the GPU. This overlap prepares
the next block's candidates while the current block is being
denoised; the next block begins once both streams complete
their work for the transition.

\para{Main stream}
The first step (\bcircle{1}) runs the model forward over the KV cache in the
candidate KV buffer.
In this forward, its $B$ real queries select the top-$K$ indices for each
KV head, and these queries are retained for merging in the retrieval
stream (\wcircle{6}).
The selection (\bcircle{2}) copies the selected critical KV subset from the candidate
buffer into the top-$K$ KV buffer.
Steps $2$ to $T$ (\bcircle{3}) run the model forward over the top-$K$
buffer until block $N$ is fully denoised.
Once every token of block $N$ is generated, a final forward computes the KV
cache of block $N$ (\bcircle{4}) over the candidate KV buffer, and this KV
cache is written to the
candidate KV buffer on the GPU and to the full KV pool in CPU DRAM
(\bcircle{5}).

\para{Retrieval stream}
\label{sec:herald:retrieval-execution}
The merge (\wcircle{6}) averages the first-step queries of block $N$, computed
by the main stream in \bcircle{1}, into one query per layer.
With this merged query, the CPU computes the scores $\mathbf{q}K^{\top}$ over
the full KV cache (\wcircle{7}).
The gather (\wcircle{8}) selects the $rK{-}B$ highest-scoring candidates and
packs the candidates into a contiguous staging buffer on the CPU.
The transfer (\wcircle{9}) moves the candidates over PCIe into the candidate
buffer on the GPU.
These steps are pipelined layer by layer, so the computation on the CPU
overlaps the transfer over PCIe.

\para{Optimized CPU Stage~1 kernel}
\label{sec:herald:cpu-kernel}
Stage~1 scans the full KV cache with a single merged query
per query head for every block and is bound by CPU memory bandwidth.
We fuse scoring, candidate selection, and gathering into one
kernel per layer.
Under GQA, the $G$ query heads sharing a KV head must produce
one ranking; separate selection passes would multiply key traffic.
For each (request, KV head), the kernel scans prefix chunks
in parallel, computing scores for all $G$ heads in one pass
and merging per-tile softmax statistics into the maximum $m_g$
and normalizer $\ell_g$ of each head~\cite{dao2022flashattentionfastmemoryefficientexact}.
The stored scores require $GS$ floats per KV head, where $S$
is the prefix length, occupying only $1/16$ of the key bytes;
writing and re-reading them therefore adds little traffic.
The kernel aggregates the heads into one importance per position,
$\hat{s}_h[s] = \sum_g \exp(\mathbf{s}[g,s]-m_g)/\ell_g$,
uses chunk-parallel radix selection to extract the
top-$(rK{-}B)$ positions, and gathers their keys and values
into a contiguous pinned buffer for GPU transfer.
Transferring the current layer's candidates overlaps with
executing the next layer's Stage~1 kernel, reducing retrieval
overhead under KV offloading.

\section{Implementation}\label{sec:impl}

We implement \name on SGLang's block-diffusion inference
path~\cite{zheng2024sglangefficientexecutionstructured}.
The main stream reuses its decoding pipeline, paged KV cache
management~\cite{kwon2023efficientmemorymanagementlarge},
and CUDA graph launches for low-overhead kernel dispatch.
All GPU attention uses FlashInfer's fused prefill and decode
kernels~\cite{ye2025flashinferefficientcustomizableattention},
which support variable-length sequences and paged KV layouts,
providing near-peak memory bandwidth utilization during
denoising without custom CUDA kernels.
The retrieval stream runs as an asynchronous background worker,
sharing the main stream's GPU context through CUDA streams.
Its only GPU operations---query merging and candidate
transfers---overlap with the main stream.
The CPU Stage~1 kernel fuses scoring, top-$(rK{-}B)$ selection,
and gathering in C++ with AVX-512 intrinsics, integrated through
a custom PyTorch extension.
CPU--GPU communication uses pinned buffers and asynchronous DMA.
The candidate and top-$K$ KV buffers are allocated at initialization
and reused across blocks, avoiding dynamic allocation during decoding.
The implementation adds approximately 3,700 lines of C++ and
9,600 lines of Python to SGLang.

\section{Evaluation}\label{sec:eval}

\subsection{Experimental Setup}\label{sec:setup}

\para{Hardware}
Performance experiments run on a cloud VM with 176 vCPUs on a two-socket
AMD EPYC 9654 host, 1.3\,TB of DDR5 DRAM with a measured STREAM bandwidth of
$221$\,GB/s, and a single NVIDIA H100 SXM 80\,GB GPU with $3.35$\,TB/s of
HBM3 bandwidth, connected via PCIe 5.0 $\times$16 with measured
host-to-device and device-to-host bandwidths of $57.6$ and $53.8$\,GB/s.

\begin{figure*}[t]
  \centering
  \includegraphics[width=\textwidth]{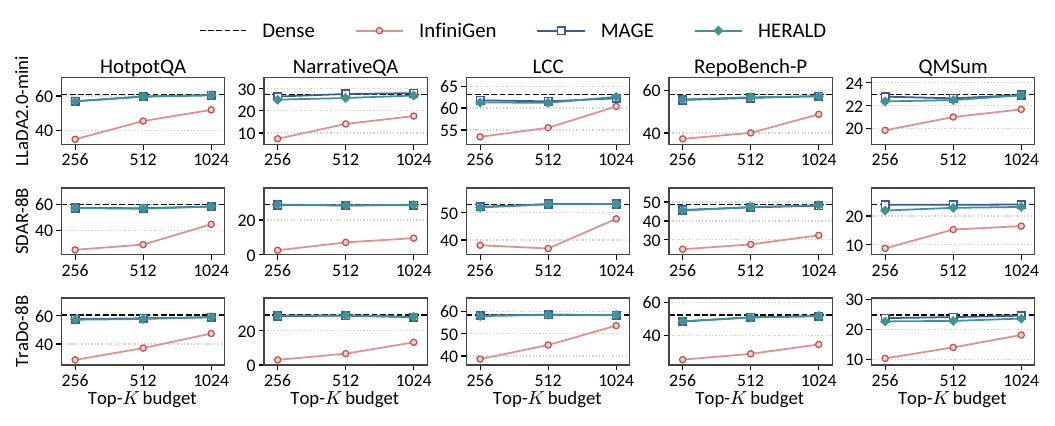}
  \caption{LongBench accuracy across all three models and five tasks at
  top-$K$ budgets of $K{=}256$, $512$, and $1024$ tokens. \name ($r{=}2$)
  closely tracks MAGE. InfiniGen, adapted from AR offloading, falls far
  below at every budget. Tables~\ref{tab:acc-raw}--\ref{tab:acc-raw-trado} in the appendix list the
  raw results.}
  \label{fig:acc-longbench}
  \vspace{-0.3cm}
\end{figure*}

\para{Models}
We evaluate four block diffusion LLMs:
SDAR-8B-Chat~\cite{cheng2025sdarsynergisticdiffusionautoregressionparadigm},
TraDo-8B-Instruct~\cite{trado},
LLaDA2.0-mini~\cite{bie2025llada20scalingdiffusionlanguage}
(16B MoE, 1.4B active), and
DreamReasoner-8B~\cite{wu2026dreamreasoner8bblocksizecurriculumlearning}.
SDAR-8B-Chat, TraDo-8B-Instruct, and DreamReasoner-8B are all adapted from Qwen3-8B and share an identical model architecture.
With a fixed number of denoising steps, throughput and latency depend only on the architecture, so we report them for SDAR-8B-Chat as a representative of the three models, together with LLaDA2.0-mini.

\para{Baselines}
All methods are built on SGLang~\cite{zheng2024sglangefficientexecutionstructured} with
FlashInfer~\cite{ye2025flashinferefficientcustomizableattention} kernels.
\textit{Dense} runs full attention with the entire KV cache in GPU memory.
\textit{MAGE}~\cite{kwon2026mage} is a sparse attention method for block
dLLMs that keeps the entire KV cache in GPU memory.
We evaluate two offloading versions of MAGE, which keep the full KV cache
in CPU memory and differ only in where the selection runs
(Section~\ref{sec:motivation:problem}).
\textit{MAGE-Offload (GPU)} streams the full KV cache of each layer to
the GPU for selection, and \textit{MAGE-Offload (CPU)} selects on the CPU
and transfers only the selected critical KV subset.
Both select exactly the same critical KV subset as MAGE, so their accuracy equals
MAGE's, and we report MAGE once in the accuracy evaluation.
\textit{InfiniGen}~\cite{lee2024infinigenefficientgenerativeinference}
is a sparse attention method for AR LLMs that predicts the critical KV
subset from the previous layer; we adapt InfiniGen to block dLLMs by
selecting with the mean query of the block.
\textit{MaskKV}~\cite{huang2025maskkv} is a KV eviction method for
diffusion LLMs; we apply it to long-generation tasks under the same
KV budget~$K$, permanently evicting entries beyond the budget.
All methods use the same KV budget, allocated uniformly across layers and
KV heads; the first two layers use full attention in all sparse methods.

\para{Configuration}
Following the conventional block dLLM setting~\cite{bie2025llada20scalingdiffusionlanguage,
wu2025fastdllmv2efficientblockdiffusion,
wu2026dreamreasoner8bblocksizecurriculumlearning}, we use
block size $B{=}32$ and confidence-based dynamic unmasking with threshold $0.95$
for all experiments.
\name uses a candidate factor of $r{=}2$ as default, and
Section~\ref{sec:eval:accuracy} evaluates how $r$ affects the
accuracy.
For LongBench tasks, we use greedy decoding, truncate prompts in the
middle to $31.5$K tokens, and cap the generation length at $512$ tokens
(i.e., $16$ blocks).
For long-generation tasks, we enable the thinking mode, sample with temperature $0.6$ and top-$p$ $0.95$, report the mean and standard deviation over three seeds, and cap
the generation length at $16$K tokens.
For \textit{throughput and latency} experiments, we fix $T$ to the average number of steps observed on LongBench ($T{=}20$) across all context lengths, use an SGLang \texttt{mem-fraction-static} of $0.85$ for all methods, and sweep batch sizes in powers of two.

\subsection{Accuracy Evaluation}\label{sec:eval:accuracy}

\begin{table}[t]
  \centering
  \scriptsize
  \setlength{\tabcolsep}{2pt}
  \caption{Accuracy of DreamReasoner-8B on long-generation tasks (thinking on,
  up to 16K generated tokens): MATH-500 accuracy and LiveCodeBench pass@1, in \%,
  mean $\pm$ standard deviation over three sampling seeds. \name uses $r{=}2$ as default; $r$ is its candidate factor.}
  \label{tab:acc-reasoning}
  \begin{tabular}{lrrrrrrr}
    \toprule
     & & & & & \multicolumn{3}{c}{\name} \\
    \cmidrule(l){6-8}
    $K$ & Dense & MaskKV & InfiniGen & MAGE & $r{=}1.5$ & $r{=}2$ & $r{=}3$ \\
    \midrule
    \multicolumn{8}{l}{\textit{MATH-500}} \\
    $512$ & \multirow{2}{*}{88.3{\tiny$\pm$1.3}} & 75.9{\tiny$\pm$1.1} & 82.9{\tiny$\pm$4.3} & 88.4{\tiny$\pm$1.9} & 85.6{\tiny$\pm$0.7} & 87.1{\tiny$\pm$1.0} & 88.5{\tiny$\pm$1.1} \\
    $1024$ & & 83.7{\tiny$\pm$0.4} & 84.6{\tiny$\pm$1.1} & 88.6{\tiny$\pm$0.8} & 86.3{\tiny$\pm$1.0} & 89.6{\tiny$\pm$0.3} & 89.3{\tiny$\pm$0.8} \\
    \midrule
    \multicolumn{8}{l}{\textit{LiveCodeBench}} \\
    $512$ & \multirow{2}{*}{39.8{\tiny$\pm$1.2}} & 15.0{\tiny$\pm$0.7} & 22.8{\tiny$\pm$7.8} & 37.2{\tiny$\pm$4.2} & 34.9{\tiny$\pm$1.5} & 35.6{\tiny$\pm$0.5} & 36.1{\tiny$\pm$2.0} \\
    $1024$ & & 23.9{\tiny$\pm$2.1} & 31.9{\tiny$\pm$1.6} & 38.1{\tiny$\pm$1.0} & 35.0{\tiny$\pm$2.2} & 40.4{\tiny$\pm$2.6} & 40.1{\tiny$\pm$0.7} \\
    \bottomrule
  \end{tabular}
\end{table}

\para{LongBench}
We evaluate generation quality on five LongBench~\cite{bai2024longbenchbilingualmultitaskbenchmark} tasks:
HotpotQA (multi-document QA), NarrativeQA (single-document QA),
QMSum (summarization), and LCC and RepoBench-P (code completion), using all
samples of each task.
Figure~\ref{fig:acc-longbench} shows the accuracy at top-$K$ budgets of
$K{=}256$, $512$, and $1024$ tokens per layer and KV head.
Across the three models, the five-task average of \name differs from MAGE's
by at most $0.4$ points at $K{=}512$ and $0.2$ points at $K{=}1024$, leaving
\name within $1.4$ and $1.0$ points of Dense.
$r{=}3$ changes the average by at most $0.5$ points
(Tables~\ref{tab:acc-raw}--\ref{tab:acc-raw-trado}).
InfiniGen loses $11$--$20$ points relative to Dense at $K{=}512$ and
$6$--$13$ points at $K{=}1024$.

\begin{figure*}[t]
  \centering
  \includegraphics[width=\textwidth]{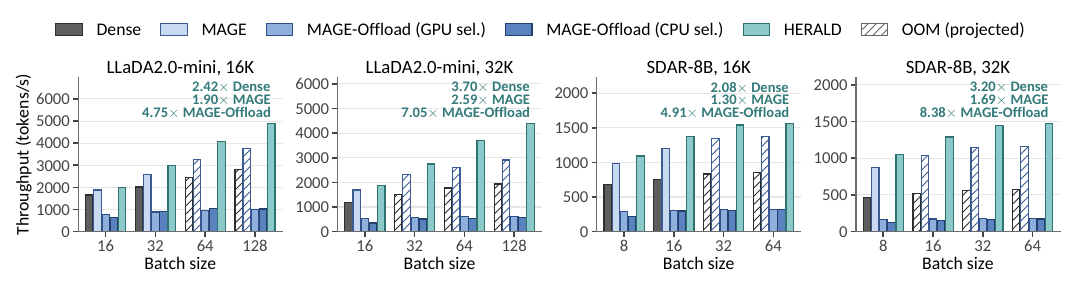}
  \caption{Decode throughput across batch sizes at 16K and 32K context
  lengths ($T{=}20$, $K{=}1024$). Dense and MAGE run out of GPU memory at
  large batches; hatched bars give their projected throughput with
  unbounded GPU memory (Figure~\ref{fig:motivation-ceiling-a}). The label
  gives \name's peak throughput relative to those of Dense,
  MAGE, and the faster MAGE-Offload variant, where each peak is taken over all batch sizes.}
  \label{fig:throughput}
\vspace{-0.3cm}
\end{figure*}

\para{Long-generation tasks}
We also evaluate DreamReasoner-8B on math and code reasoning tasks
using MATH-500~\cite{math500} and LiveCodeBench~\cite{lcb}, respectively,
where a short prompt is followed by thousands of generated tokens across
many decode blocks (Table~\ref{tab:acc-reasoning}).
At $K{=}1024$, \name matches Dense and MAGE on both benchmarks;
at $K{=}512$, it stays within 1.6 percentage points of MAGE.
Compared with the default $r{=}2$, increasing the candidate factor to
$r{=}3$ improves accuracy by at most 1.4 points at $K{=}512$, while
reducing it to $r{=}1.5$ lowers accuracy by up to 5.4 points across the
two budgets, illustrating the tradeoff between accuracy and
candidate-buffer size.
InfiniGen again falls below Dense on both benchmarks.
For these long-generation tasks, we additionally evaluate MaskKV as a
representative KV eviction method under the same KV budgets.
KV eviction permanently discards entries that cannot be recovered when
later blocks need them. MaskKV achieves the lowest accuracy among the
evaluated methods on both benchmarks, trailing Dense by 4.6--12.4
percentage points on MATH-500 and 15.9--24.9 points on LiveCodeBench.

In summary, at $K{=}1024$, where MAGE approaches Dense, \name with
$r{=}2$ also matches Dense across all models and tasks.
InfiniGen and MaskKV fall far below. We therefore exclude both methods
from the throughput evaluation.

\subsection{Performance Evaluation}

\begin{figure}[t]
  \centering
  \includegraphics[width=\columnwidth]{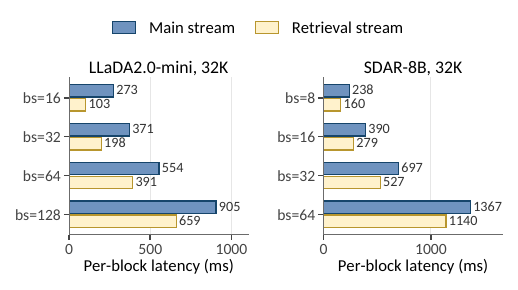}
  \caption{Per-block latency of \name's main stream and retrieval stream at 32K
  context ($K{=}1024$, $T{=}20$).}
  \label{fig:latency-breakdown}
    \vspace{-0.3cm}
\end{figure}

We evaluate the decode throughput and the per-block latency breakdown of \name
against Dense, MAGE, and MAGE-Offload with GPU and CPU selection at 16K and 32K
context lengths. All sparse methods use the same top-$K$ budget of $K{=}1024$
tokens per layer and KV head.

\para{Throughput Evaluation}
Figure~\ref{fig:throughput} shows the decode throughput across batch sizes.
Dense and MAGE keep the full KV cache in GPU memory and run out of memory
beyond a batch size of 32 at 16K and 16 at 32K on LLaDA2.0-mini, and 16
and 8 on SDAR-8B.
MAGE-Offload sustains larger batches, but the critical KV selection and
the commit over the full KV cache at every block limit its throughput to
0.27--0.52$\times$ the peak of Dense.
In contrast, \name reaches 2.42$\times$ and 3.70$\times$ the peak
throughput of Dense at 16K and 32K on LLaDA2.0-mini, and 2.08$\times$
and 3.20$\times$ on SDAR-8B. It also achieves up to 8.38$\times$ the peak
throughput of the faster MAGE-Offload variant.
\name also reaches 1.30--2.59$\times$ the peak of MAGE and even exceeds
MAGE's projected throughput with unbounded GPU memory, because its
first-step selection and commit attend only to the candidate KV buffer
rather than the full KV cache.

\para{Overlap of the main and retrieval streams}
\label{sec:breakdown}
Figure~\ref{fig:latency-breakdown} shows the per-block latency of
\name's main stream and retrieval stream.
At every batch size, the retrieval stream finishes Stage~1 for the next
block while the main stream runs the $T$ denoising steps of the current
block, so \name hides the KV-retrieval overhead behind denoising.
The retrieval stream ends 62\% and 27\% of the block earlier than the main
stream at batch sizes of 16 and 128 on LLaDA2.0-mini, and 33\% and 17\%
earlier at 8 and 64 on SDAR-8B.

\section{Conclusion}
\label{sec:conclusion}
This paper presented \name, a dynamic sparse attention method for
block dLLMs that offloads the full KV cache to CPU DRAM.
Its two-stage selection uses a merged query from the previous block
to search for candidates on the CPU, then re-ranks these candidates
with the current block's real queries on the GPU to select the final
critical KV subset. This design reduces selection cost and overlaps
CPU candidate search with GPU denoising.
\name achieves near-lossless accuracy relative to Dense on LongBench
and reasoning tasks, with up to 3.70$\times$ the decode throughput
of GPU-only inference and 8.38$\times$ that of MAGE-Offload.

\bibliography{refs}
\bibliographystyle{mlsys2025}
\section*{AI Usage Disclosure}
We used generative AI tools in the following ways.
\textbf{Writing:} We used Claude (Anthropic) and OpenAI Codex
to refine phrasing, correct grammar, and improve clarity.
The authors reviewed the resulting text and checked every
technical claim and reported numerical result.
\textbf{Code and figures:} Claude Code and OpenAI Codex assisted
with supporting code, including data-processing and plotting
scripts, and with figure preparation and layout.
\textbf{Author contributions:} The authors developed HERALD's
two-stage KV selection algorithm and CPU--GPU execution system,
designed and conducted the experiments, and analyzed and
validated the results.

\ifwithappendix
\clearpage
\appendix
\section{Roofline of CPU Selection}
\label{app:roofline}

\begin{figure}[!h]
  \centering
  \includegraphics[width=\columnwidth]{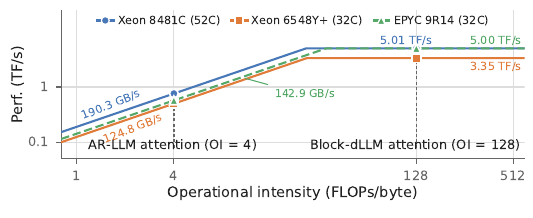}
  \caption{Roofline analysis of AR-LLM and block-dLLM attention on three
  CPU platforms. The roofs are the measured DRAM bandwidth (GB/s) and the
  measured BF16{+}FMA compute peak (TF/s), the harmonic mean of the
  AVX-512 BF16-dot and FP32-FMA peaks that exact attention uses. Both
  evaluated models, LLaDA2.0-mini and SDAR-8B-Chat ($G{=}4$), share the two
  workload positions.}
  \label{fig:motivation-roofline}
\end{figure}

Figure~\ref{fig:motivation-roofline} places exact attention on the
rooflines of three CPU platforms. Its operational intensity equals the
number of query tokens that share one read of the KV cache. A single AR
query ($G{=}4$ with GQA) stays below every ridge point (26--35
FLOPs/byte), whereas a block dLLM's $B{=}32$ \texttt{[MASK]} queries raise
the intensity to $BG{=}128$, beyond every ridge point, so the CPU's compute
caps the selection (Section~\ref{sec:motivation:problem}).

\section{Accuracy Results}
\label{app:acc-raw}

Tables~\ref{tab:acc-raw}--\ref{tab:acc-raw-trado} list the LongBench scores plotted in
Figure~\ref{fig:acc-longbench} for every model, task, top-$K$ budget, and method.

\begin{table}[!h]
  \centering
  \caption{LongBench accuracy of LLaDA2.0-mini. HQA: HotpotQA; NQA: NarrativeQA; RB: RepoBench-P; QM: QMSum. $K$ is the token budget per layer and KV head. Avg.\ is the five-task mean, with its difference from Dense in parentheses. \name defaults to $r{=}2$; $r{=}3$ is shown for comparison.}
  \label{tab:acc-raw}
  \fontsize{7}{8.5}\selectfont
  \setlength{\tabcolsep}{2pt}
  \begin{tabular}{@{}lrrrrrr@{}}
    \toprule
    Method & HQA & NQA & LCC & RB & QM & Avg. \\
    \midrule
    Dense & 60.89 & 27.32 & 63.13 & 58.00 & 22.98 & \textbf{46.46} \\
    \cmidrule(lr){1-7}
    \multicolumn{7}{c}{$K=256$} \\
    InfiniGen & 34.87 & 7.35 & 53.41 & 37.21 & 19.83 & \textbf{30.53} ($-15.93$) \\
    MAGE & 56.93 & 26.39 & 61.79 & 55.45 & 22.79 & \textbf{44.67} ($-1.79$) \\
    \name ($r{=}2$) & 56.93 & 24.99 & 61.30 & 55.68 & 22.37 & \textbf{44.25} ($-2.21$) \\
    \name ($r{=}3$) & 56.93 & 25.29 & 61.92 & 55.48 & 22.66 & \textbf{44.46} ($-2.01$) \\
    \cmidrule(lr){1-7}
    \multicolumn{7}{c}{$K=512$} \\
    InfiniGen & 45.46 & 14.05 & 55.49 & 40.00 & 21.00 & \textbf{35.20} ($-11.26$) \\
    MAGE & 59.67 & 27.55 & 61.55 & 56.37 & 22.64 & \textbf{45.56} ($-0.91$) \\
    \name ($r{=}2$) & 59.68 & 25.73 & 61.22 & 56.70 & 22.49 & \textbf{45.16} ($-1.30$) \\
    \name ($r{=}3$) & 59.76 & 27.09 & 61.59 & 57.01 & 22.89 & \textbf{45.67} ($-0.80$) \\
    \cmidrule(lr){1-7}
    \multicolumn{7}{c}{$K=1024$} \\
    InfiniGen & 51.94 & 17.59 & 60.38 & 48.72 & 21.67 & \textbf{40.06} ($-6.40$) \\
    MAGE & 60.49 & 27.91 & 62.25 & 57.23 & 22.93 & \textbf{46.16} ($-0.30$) \\
    \name ($r{=}2$) & 60.48 & 26.85 & 62.58 & 57.13 & 22.90 & \textbf{45.99} ($-0.48$) \\
    \name ($r{=}3$) & 60.41 & 26.41 & 62.49 & 57.27 & 22.97 & \textbf{45.91} ($-0.55$) \\
    \bottomrule
  \end{tabular}
\end{table}

\begin{table}[!h]
  \centering
  \caption{LongBench accuracy of SDAR-8B-Chat. Notation follows Table~\ref{tab:acc-raw}.}
  \label{tab:acc-raw-sdar}
  \fontsize{7}{8.5}\selectfont
  \setlength{\tabcolsep}{2pt}
  \begin{tabular}{@{}lrrrrrr@{}}
    \toprule
    Method & HQA & NQA & LCC & RB & QM & Avg. \\
    \midrule
    Dense & 59.83 & 28.93 & 52.98 & 48.62 & 23.97 & \textbf{42.87} \\
    \cmidrule(lr){1-7}
    \multicolumn{7}{c}{$K=256$} \\
    InfiniGen & 25.18 & 2.56 & 38.07 & 25.01 & 8.70 & \textbf{19.90} ($-22.96$) \\
    MAGE & 57.20 & 28.74 & 52.16 & 45.80 & 23.94 & \textbf{41.57} ($-1.30$) \\
    \name ($r{=}2$) & 57.25 & 28.99 & 51.95 & 45.52 & 21.92 & \textbf{41.13} ($-1.74$) \\
    \name ($r{=}3$) & 57.24 & 29.03 & 51.89 & 45.55 & 22.44 & \textbf{41.23} ($-1.64$) \\
    \cmidrule(lr){1-7}
    \multicolumn{7}{c}{$K=512$} \\
    InfiniGen & 29.20 & 7.12 & 36.87 & 27.59 & 15.27 & \textbf{23.21} ($-19.66$) \\
    MAGE & 57.13 & 28.45 & 53.19 & 47.18 & 23.94 & \textbf{41.98} ($-0.89$) \\
    \name ($r{=}2$) & 56.68 & 28.51 & 53.08 & 47.35 & 22.85 & \textbf{41.69} ($-1.17$) \\
    \name ($r{=}3$) & 56.68 & 28.55 & 52.90 & 47.06 & 22.98 & \textbf{41.63} ($-1.23$) \\
    \cmidrule(lr){1-7}
    \multicolumn{7}{c}{$K=1024$} \\
    InfiniGen & 44.77 & 9.57 & 47.78 & 32.39 & 16.46 & \textbf{30.19} ($-12.67$) \\
    MAGE & 58.41 & 28.67 & 53.19 & 48.05 & 24.00 & \textbf{42.46} ($-0.40$) \\
    \name ($r{=}2$) & 58.41 & 28.83 & 53.22 & 48.33 & 23.16 & \textbf{42.39} ($-0.48$) \\
    \name ($r{=}3$) & 58.41 & 28.83 & 53.26 & 48.42 & 23.25 & \textbf{42.43} ($-0.43$) \\
    \bottomrule
  \end{tabular}
\end{table}

\begin{table}[!h]
  \centering
  \caption{LongBench accuracy of TraDo-8B-Instruct. Notation follows Table~\ref{tab:acc-raw}.}
  \label{tab:acc-raw-trado}
  \fontsize{7}{8.5}\selectfont
  \setlength{\tabcolsep}{2pt}
  \begin{tabular}{@{}lrrrrrr@{}}
    \toprule
    Method & HQA & NQA & LCC & RB & QM & Avg. \\
    \midrule
    Dense & 60.55 & 29.20 & 58.42 & 52.48 & 24.78 & \textbf{45.09} \\
    \cmidrule(lr){1-7}
    \multicolumn{7}{c}{$K=256$} \\
    InfiniGen & 28.82 & 2.92 & 38.63 & 25.26 & 10.26 & \textbf{21.18} ($-23.91$) \\
    MAGE & 57.54 & 28.48 & 58.11 & 48.41 & 23.68 & \textbf{43.24} ($-1.84$) \\
    \name ($r{=}2$) & 57.10 & 28.77 & 57.83 & 48.55 & 22.57 & \textbf{42.96} ($-2.12$) \\
    \name ($r{=}3$) & 57.07 & 28.64 & 58.05 & 48.34 & 23.00 & \textbf{43.02} ($-2.07$) \\
    \cmidrule(lr){1-7}
    \multicolumn{7}{c}{$K=512$} \\
    InfiniGen & 37.11 & 6.54 & 44.93 & 28.78 & 13.93 & \textbf{26.26} ($-18.83$) \\
    MAGE & 58.10 & 28.66 & 58.47 & 50.98 & 24.01 & \textbf{44.04} ($-1.04$) \\
    \name ($r{=}2$) & 57.60 & 28.85 & 58.43 & 50.84 & 22.75 & \textbf{43.69} ($-1.39$) \\
    \name ($r{=}3$) & 58.07 & 28.68 & 58.63 & 50.93 & 22.93 & \textbf{43.85} ($-1.24$) \\
    \cmidrule(lr){1-7}
    \multicolumn{7}{c}{$K=1024$} \\
    InfiniGen & 47.39 & 13.15 & 53.59 & 34.47 & 18.05 & \textbf{33.33} ($-11.76$) \\
    MAGE & 58.84 & 27.86 & 58.23 & 51.67 & 24.49 & \textbf{44.22} ($-0.87$) \\
    \name ($r{=}2$) & 58.89 & 28.05 & 58.35 & 51.81 & 23.53 & \textbf{44.13} ($-0.96$) \\
    \name ($r{=}3$) & 58.84 & 27.96 & 57.85 & 51.78 & 23.30 & \textbf{43.95} ($-1.14$) \\
    \bottomrule
  \end{tabular}
\end{table}

\fi

\end{document}